\documentclass[10pt,conference,letterpaper]{ieeeconf}

\usepackage{hyperref}
\usepackage{times}
\usepackage{epsfig}
\usepackage{graphicx}
\usepackage{amsmath}
\usepackage{amssymb}
\usepackage{booktabs}
\graphicspath{ {./images/} }
\usepackage{svg}

\newcommand{\etal}{\textit{et al}. }

\begin{document}

\title{Under pressure: learning-based analog gauge reading in the wild}

\author{Maurits Reitsma$^{1}$, Julian Keller$^{1}$, Kenneth Blomqvist$^{1}$ and Roland Siegwart$^{1}$ \\
$^{1}$Autonomous Systems Lab \\ 
ETH Z\"{u}rich, Switzerland
}

\maketitle
\thispagestyle{empty}

\begin{abstract}
We propose an interpretable framework for reading analog gauges that is deployable on real world robotic systems. Our framework splits the reading task into distinct steps, such that we can detect potential failures at each step. Our system needs no prior knowledge of the type of gauge or the range of the scale and is able to extract the units used. We show that our gauge reading algorithm is able to extract readings with a relative reading error of less than 2\%.
\end{abstract}

\section{Introduction}
Robots are often tasked to inspect industrial plants. One critical part of inspecting industrial plants is reading analog gauges. While digital gauges exist, the vast majority of gauges are still analog. Analog gauges are designed to be easy for humans to read, but they are challenging to read using state-of-the-art computer vision algorithms. There are a plethora of different gauge designs. They vary in units and measurement scales. The gauge face is typically protected with glass, which shows reflections and can be covered with dirt. The gauges can be mounted in tricky positions which are impossible for the robot to approach and perceive head on, so reading needs to be possible from a variety of different orientations and angles. To build robots that can inspect and interact with legacy infrastructure, we need to develop algorithms that can reliably read analog gauges.

In this paper, we present a learning-based system to autonomously read analog industrial gauges \cite{analog_gauge_reader}. Our system aims to achieve robust performance across gauges found in real-world scenarios with diverse appearance, scales and units. We do not require prior information about the gauge, ensuring broad applicability in industrial environments. 

Our approach builds on the insight that all analog gauges we have seen have a face for the dial, a needle and a scale with notches along the way. We first detect and segment gauge faces and needles in an image. For each detection, we apply a keypoint detection algorithm to detect major notches on the scale of the gauge. As there can be a perspective shift and most gauges don't actually have a circular scale or they have a scale with a larger radius than the gauge face, we fit an ellipse to the detected keypoints. We then run optical character recognition (OCR) on the gauge face to detect units and scale markings on the gauge face, which we anchor to the ellipse. We fit a linear model with a RANSAC loop to the detected markings, as some might have been misread. We finally find the intersection point between the needle and the estimated ellipse and predict the gauge reading by interpolation using our linear model.

We evaluate our algorithm on a set of analog gauges collected from the web, as well as a proprietary dataset of high quality pressure gauge images collected in an operating oil refinery. Additionally, we evaluate our system on the dataset provided by Howells \etal \cite{realTimeGaugeTranscription}.

\begin{figure}[t]
\begin{center}
\includegraphics[width=\linewidth]{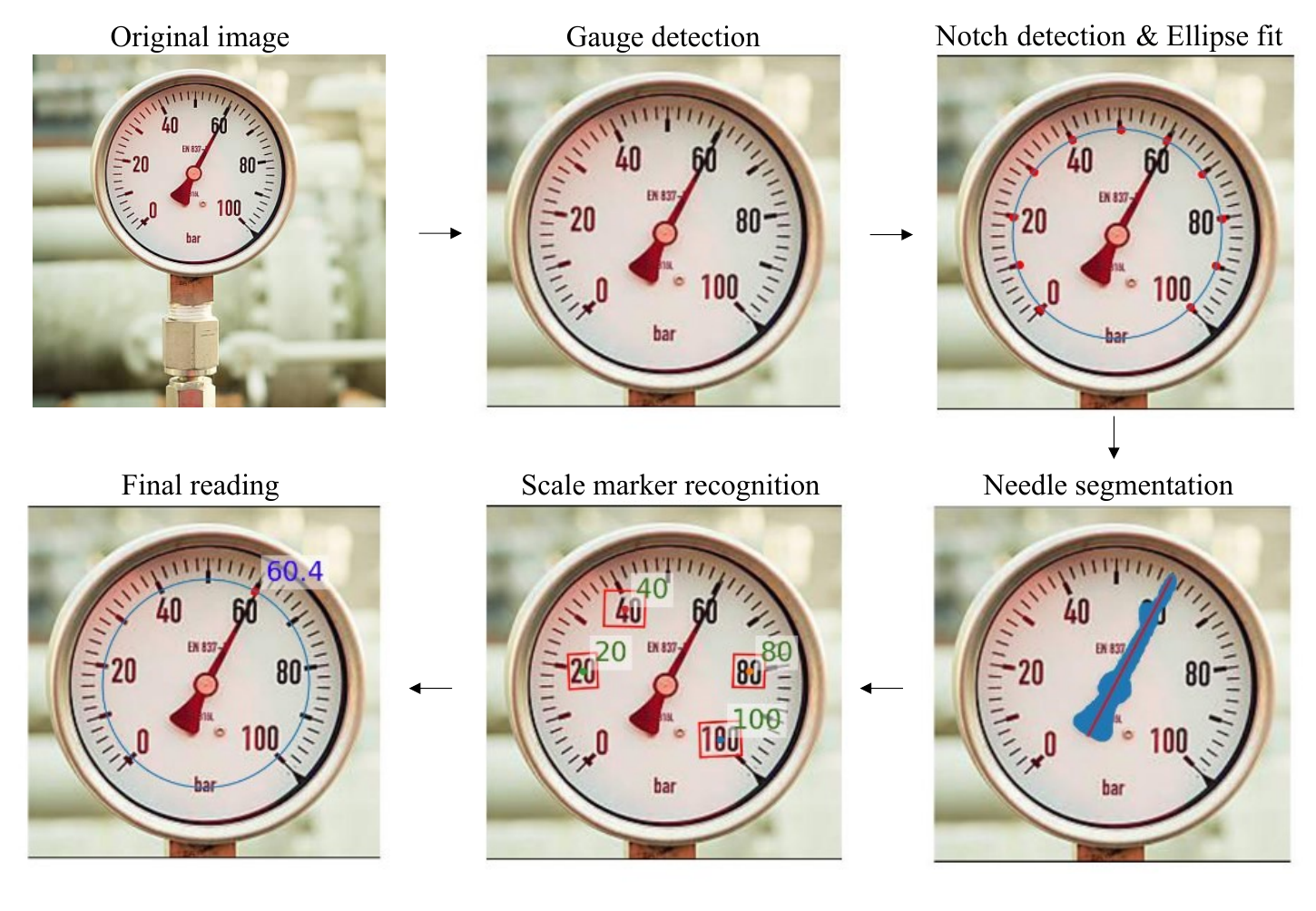}
\end{center}
\vspace{-0.5cm}
\caption{Method overview. We first detect the gauge in the image, followed by detecting the notches and fitting an ellipse through them. Then, we segment the needle and fit a line through it. We detect and read the characters of the scale markers. Finally, we use the extracted information to compute the final reading.}
\vspace{-0.5cm}
\label{fig:short}
\label{fig:method_overview}
\end{figure}

We show that our method yields a relative reading error of around $2\%$, beating the work of Howells et al. \cite{realTimeGaugeTranscription} on their dataset, while not requiring any prior gauge information, showcasing the potential of our algorithm. We analyze failure cases of our algorithm and show how our system may be improved in the future. We release our software as open source to allow future engineers and researchers to leverage our work and make further progress on the problem. \footnote{\href{https://github.com/ethz-asl/analog_gauge_reader}{https://github.com/ethz-asl/analog\_gauge\_reader}}

\section{Related Work}
Several works in recent years have tackled autonomous analog gauge reading. Localizing the gauge is generally the first stage in a pipeline for reading gauges in recent works on the subject \cite{analogClockReading, realTimeGaugeTranscription, dumberger20, gauread22}, as it reduces the background noise for the further steps.

Similarly to us Howells \etal \cite{realTimeGaugeTranscription} use a learning-based approach. Their method extracts the start and end point of the scale, the center of the gauge and the needle tip to read the gauge. However, this assumes prior knowledge about the range of the scale of the gauge. We do not make this assumption which makes our algorithm more general.

Some recent works do not make assumptions about the scale of the gauge. Dumberger \etal \cite{dumberger20} incorporate an optical character recognition (OCR) step to read the scale. However, they impose additional restrictions. They assume that the gauge is not being partially covered by objects or and require the scale to be circular and spanning at least a range of 180 degrees. We avoid this making our algorithm capable of handling a wider range of gauge types and scenarios.

Another recent approach similar to ours, which incorporates an OCR stage, was proposed by Milana \etal \cite{gauread22}. However, there are some key differences between their method and ours. Milana \etal employed a technique where they warp the image, transforming the gauge face from an ellipse into a circular shape. They then directly compute the final prediction using polar coordinates on the warped image. In contrast, our approach assumes a circular shape of the gauge scale, rather than assuming the entire gauge face to be circular. This key distinction allows our method to apply to a wider range of gauge types. Furthermore, our approach offers an additional advantage when dealing with gauges that have multiple scales. Our approach allows the distinction between scale markers of the inner and outer scale of the gauge.

\section{Methodology}
We propose a robust and modular analog gauge reading algorithm with the following stages: (1) gauge detection, (2) notch detection, (3) ellipse fitting, (4) needle segmentation, (5) character recognition of scale markings and units, (6) computing reading through linear model and interpolation.

To achieve a more robust system, we leverage some assumptions. Firstly, we assume that the gauge has a linear scale with at least 5 visible major notches on the scale aligned in an arc on the gauge face. We further assume that there are numerical scale markers along the scale and that the gauge face is flat. Additionally, we assume that the point between the start and end notch of an unrotated gauge is on the bottom. An overview is shown in Figure \ref{fig:method_overview}. In the following subsections, we describe each stage in detail.

\subsection{Gauge detection}
As a first step, we detect the gauge and crop the image to separate different gauges in the image and to remove any background noise. We use the smallest off-the-shelf YOLOv8 detector \cite{YOLO_by_Ultralytics_2023} to detect the gauge. We found the smallest model to be more than sufficient for this task. The model has been pretrained on the COCO dataset \cite{Coco}, which we further fine-tune on a small dataset of around 2,000 labeled gauge images. After detection, we crop and resize the image using bilinear interpolation to a resolution of $(448, 448)$.

\subsection{Notch detection}

\begin{figure}[t]
\begin{center}
\includegraphics[width=\linewidth]{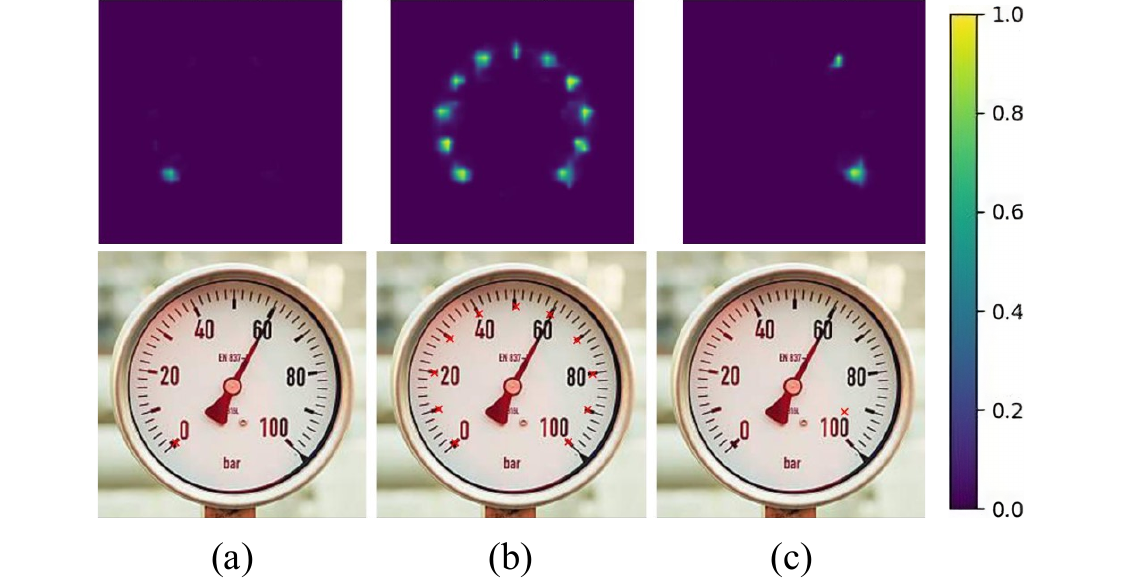}
\end{center}
\vspace{-0.4cm}
   \caption{Notch detection overview. First, we predict a heatmap to predict the probability of each pixel being a notch, then we cluster this heatmap to extract the individual notch points. In (b) we predict all notches, while in (a) and (c) we predict the start and end notch respectively.}
\vspace{-0.5cm}
\label{fig:notch_detection}
\end{figure}

After gauge detection, we detect notches on the scale. We do this by designing a simple keypoint detector. We split notch detection into three different classes of keypoints: intermediate notches, the start notch and the end notch. The start and end notches are used to fix the orientation in a later stage of the pipeline, as detailed in Section \ref{ocr}.
We formulate keypoint detection as a heatmap prediction problem, followed by clustering to separate keypoint detections from each other. Figure \ref{fig:notch_detection} shows the different steps.

\subsubsection{Keypoint prediction}

To arrive at a vision model that can be trained from a small number of examples, we use a pretrained vision transformer DINOv2 backbone \cite{dinov2}. To predict heatmaps from the extracted features, we add a decoder structure consisting of a two-stage 1x1 CNN layer network, followed by a bilinear upsampling step. The upsampling is followed by a sigmoid activation. To construct the heatmaps for training, we place a Gaussian distribution centered at each keypoint location. We normalize heatmaps to have values in the range $[0, 1]$. We predict three heatmaps. One for intermediate notches, one for the start notch and one for the end notch.

To train the model, we freeze the backbone parameters and optimize the rest using a binary cross-entropy loss function and stochastic gradient descent and the Adam optimizer \cite{adam}.

To extract keypoints from the predicted heatmaps, we use the Mean-Shift algorithm \cite{mean_shift}, clustering heatmap indices which have a value higher than $0.5$.

\subsection{Ellipse fit}

Gauge scales are typically on an arc. To approximate the arc of the scale, we fit an ellipse through the coordinates of the keypoints of the detected notches. To fit an ellipse to a set of two-dimensional data points, we use the algorithm by Halir \etal \cite{ellipse_numerically_stable}. It is a numerically stable version of the algorithm proposed by Fitzgibbon \etal \cite{ellipse}. Using direct least squares solving a corresponding eigensystem, the algorithm efficiently determines the optimal ellipse fit for the given data points. The ellipse fitting requires at least 5 notches to be detected.

\subsection{Needle segmentation and line fit}

We tackle the problem of estimating the position and direction of the gauge needle in two steps. In the first step, we segment pixels belonging to the needle using an object instance segmentation model. In the second step, we fit a line through the segmented pixels. Our approach implicitly assumes that the needle is directed towards the symmetry axis of the needle, which is a valid assumption for the vast majority of analog industrial gauges.

For the instance segmentation model, we use a COCO pretrained \cite{Coco} YOLOv8 model \cite{YOLO_by_Ultralytics_2023}. We fine-tune the model on our dataset of around 2,000 labeled images of gauges. To fit the line through the segmented needle pixels, we use Orthogonal Distance Regression \cite{odr}, as it minimizes the shortest distance of each point to the line.

\subsection{Detection of scale numbers and unit} \label{ocr}

\begin{figure}[t]
\begin{center}
\includegraphics[width=\linewidth]{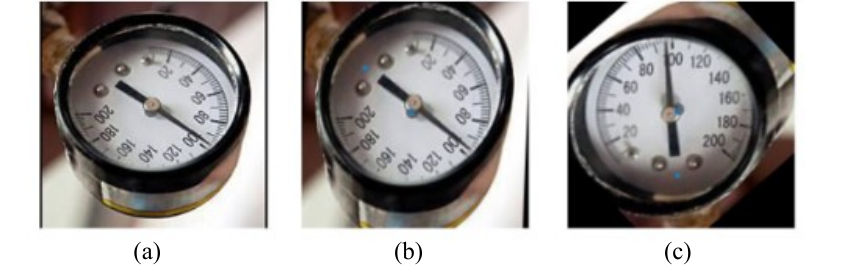}
\end{center}
\vspace{-0.4cm}
   \caption{Visualization of the perspective and rotation correction. (a) shows the original cropped image. (b) shows the image after perspective correction. (c) shows the image after rotation correction.}
\vspace{-0.5cm}
\label{fig:ocr_warp_rot}
\end{figure}

As we don't know the range of the gauge scale, we extract the numbers and unit displayed on the gauge face. We do this using an optical character recognition (OCR) model. We use a two-stage pre-trained deep neural network. The first network is dedicated to text detection, and the second network does text recognition. For text detection, we use the DBNet model \cite{DBNet}, which is a segmentation model using a ResNet-18 backbone \cite{ResNet18}. For text recognition, we use an ABINet model \cite{ABINet}. The ABINet architecture uses a combination of a vision model and a language model to predict the text in the detected bounding box. For both models, we use the MMOCR library \cite{kuang2021mmocr}.

The OCR pipeline is sensitive to rotations and perspective shifts. To correct for this, we use the detected ellipse, start notch, and end notch. To correct for perspective, we warp the gauge image, such that the computed ellipse becomes circular. We assume that we have a circular scale of some, however large, radius. We use the start and end notch to estimate the orientation of the gauge by assuming the point between the start and end notch is on the bottom of the gauge. This allows us to estimate the gauge orientation and saves computation time, compared to exhaustively sampling many rotations. Both transformations are shown in Figure \ref{fig:ocr_warp_rot}.

\subsection{Computing the final reading}

After running the previous stages of the pipeline, we compute the reading of the gauge using the following steps, which we explain in the following two subsections: (1) scale marker projection to ellipse and scale interpolation and (2) needle-to-ellipse intersection.

\subsubsection{Projection and interpolation of scale markers} \label{project_point}

From the text recognitions from the OCR stage, we consider only the subset of recognitions that can be interpreted as numerical values. For each such detection, we project the center of the corresponding bounding box to the ellipse to anchor it.

\begin{figure}[t]
\begin{center}
\includegraphics[width=0.75\linewidth]{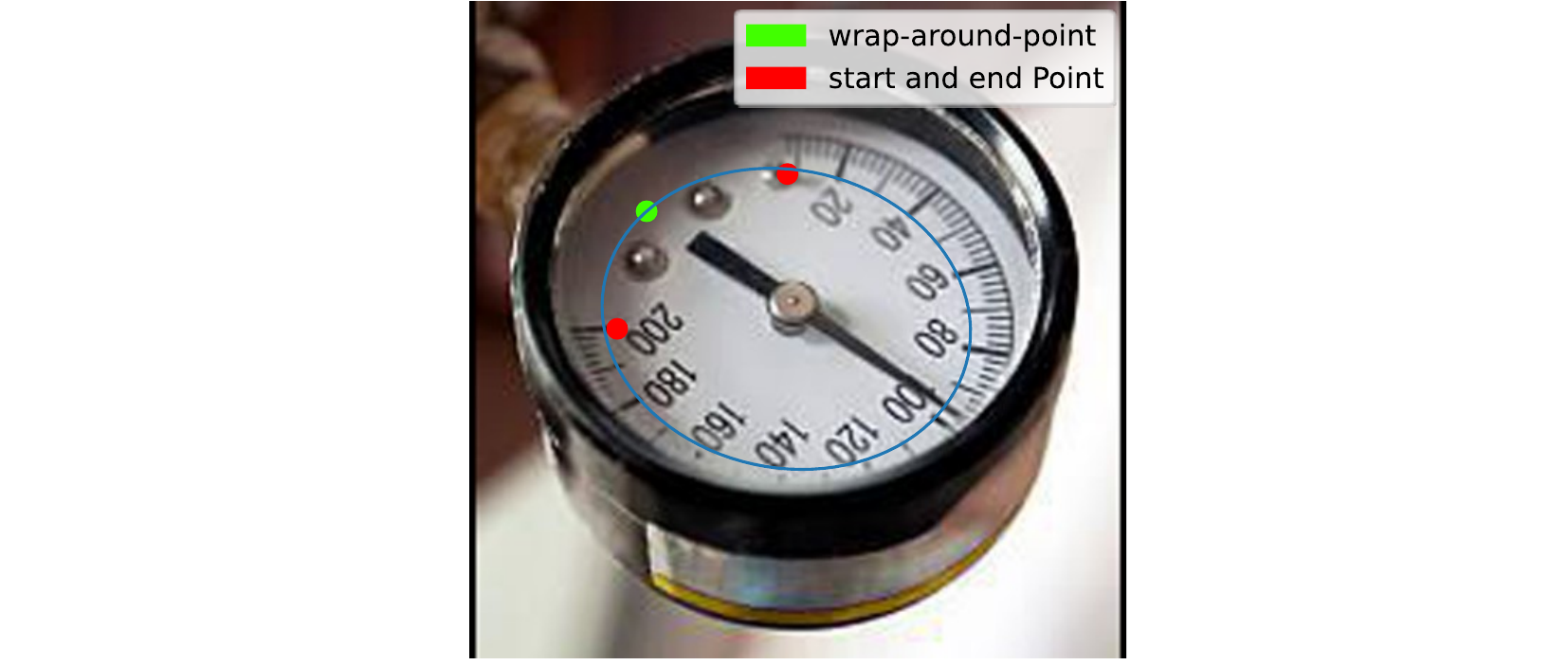}
\end{center}
\vspace{-0.4cm}
\caption{Example of wrap-around point on ellipse. We take the middle point between the start and end point. This allows us to force the wrap-around point to be outside of the scale.}
\vspace{-0.5cm}
\label{fig:ellipse_zero_point}
\end{figure}

By projecting scale markers onto the ellipse, we can determine the needle angle for each marker. Using the angles, we fit a correspondence line that maps angles on the ellipse to a reading on the scale. This can be done using least squares on the angles of the scale markers, resulting in a model which we can interpolate.

\begin{figure}[t]
\begin{center}
\includegraphics[width=\linewidth]{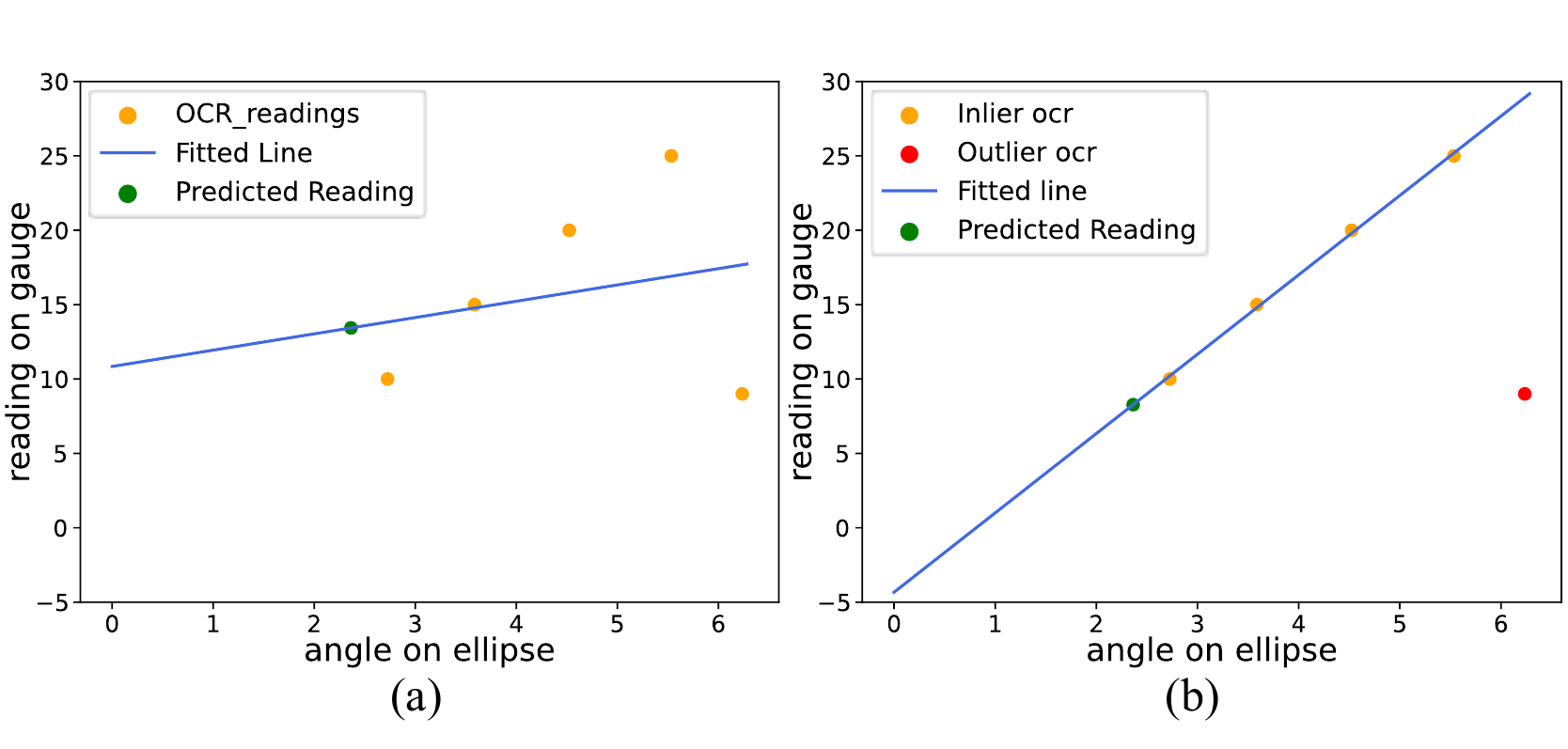}
\end{center}
\vspace{-0.5cm}
   \caption{Interpolation of the scale markers to fit the line mapping the ellipse angle to the gauge value reading. (a) Shows the least squares fit, which is heavily disturbed by outliers. (b) Shows the improved least squares fit by using RANSAC.}
\vspace{-0.5cm}
\label{fig:interpolation_ransac}
\end{figure}

However, linear regression with least squares is sensitive to outliers. Outliers can occur for many reasons, such as faulty OCR readings and detections of unrelated numbers on the gauge, such as a serial number.
We remove outliers using RANSAC \cite{RANSAC}. We fit a linear model mapping ellipse angles to value readings, filtering out any values that do not fit the model. Figure \ref{fig:interpolation_ransac} shows an example of this, where RANSAC improves the final prediction. We fit a line, as we assume that the scale is linear. In cases where the scale represents a different function, another type of mapping would have to be chosen, possibly at runtime.

To support gauges with multiple scales, we can distinguish between scale markers for the inner and outer scale, by checking if they lie inside or outside of the ellipse. By making this distinction, we can proceed to fit two linear models, one for each scale, and subsequently calculate a reading for each scale.

When mapping a linear scale using angles from $0$ to $2\pi$, the position of the zero angle becomes significant. We refer to this point as the "wrap-around point" on the ellipse. We choose the wrap-around point to be the point in between the scale start and end points, as shown in Figure \ref{fig:ellipse_zero_point}. This guarantees that the chosen wrap-around point is outside of the scale, preventing discontinuities in the line fitting.

\subsubsection{Intersection of needle and ellipse}

The final step is to determine the point on the ellipse that the needle is pointing at. We do this by finding the intersection of the needle line and the ellipse. We first rotate and shift the needle line, such that the ellipse is centered at the origin. The intersection is computed using the equation of the line $y=mx + c$ and the ellipse. By computing the roots of the following quadratic equation, we find the two intersection points:

\[ \frac{x^{2}}{a^{2}} + \frac{(mx + c)^{2}}{b^{2}} = 1\]

We then pick the intersection point that lies on our needle line. If this applies to neither or both points, we pick the one that is closer to either end of our needle line, as determined by the segmentation. This heuristic is based on the observation that the needle tip is usually closer to the gauge scale than the tail of the needle.

After determining the intersection point, we compute the angle of this point on the ellipse using the formula from Section \ref{project_point}. With the needle angle, we predict the reading using the line model mapping angles to readings.

\section{Experiments}
We test our system on two datasets. The first consists of images of many different gauge types that are crawled from the web. The other consists of proprietary gauge images collected in an oil refinery. Both sets contain a variety of different gauge types.

To assess the performance of our system, we partition our dataset into several categories. The categorization is based on the specific difficulties posed by each image, allowing us to measure how our algorithm deals with these challenges. For the first category, we select images of gauges taken from the front. The second category consists of images viewing the gauge at an angle. The third contains images where the gauge is rotated and the fourth category is for images where the gauges are both rotated and viewed at an angle. The fifth and final category contains gauges that look significantly different from the gauges in our training set, for example having a non-conventional needle type and color, or a scale covering only a small part of the gauge face while having a very large radius.

We compare the estimated reading with expert-annotated labels. For cases where a reading cannot be computed or the reading significantly deviates from the true value, we identify the stage of the pipeline responsible for the failure and examine the underlying reasons.

We perform an ablation study, to evaluate the effect of the perspective and rotation correction on the OCR stage. Furthermore, we compare our algorithm method published by Howells \etal \cite{realTimeGaugeTranscription} on their dataset. This dataset consists of 450 images each of 5 different gauge types. We excluded meter\_d because as it has a nonlinear scale.

\begin{figure}[t]
\begin{center}
\includegraphics[width=\linewidth]{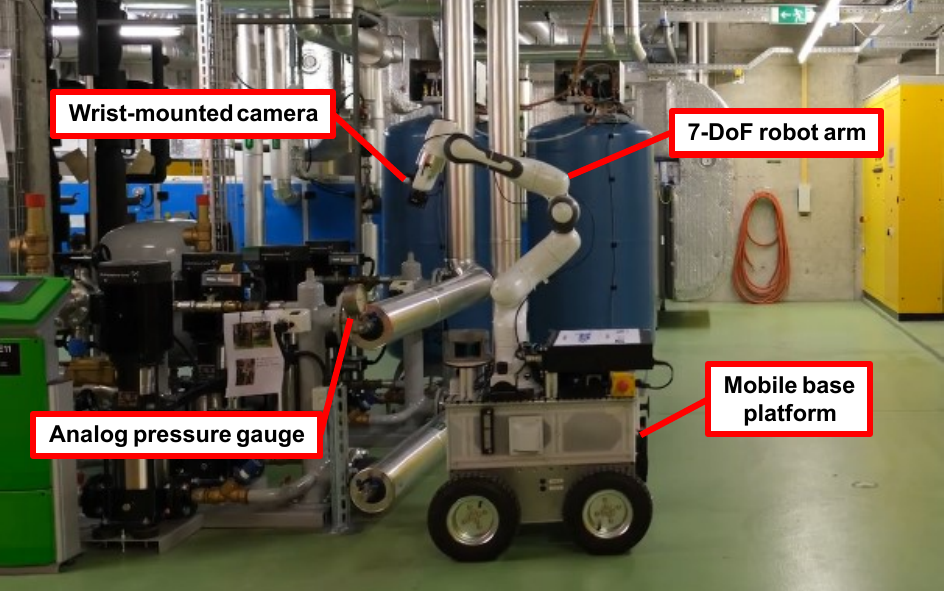}
\end{center}
\vspace{-0.4cm}
   \caption{Real-world gauge reading deployment on a mobile robotic base platform with manipulator and wrist-mounted camera.}
\label{fig:robot_setup}
\end{figure}

Next to evaluations on datasets, we deployed our system on a mobile robotic inspection and maintenance system, as shown in Figure \ref{fig:robot_setup}. Our system allows for consecutive readings of gauges at different locations and heights within industrial plants.

\section{Results}

\begin{table}
\tiny
\begin{center}
\begin{tabular}{l|cc}
\toprule
Test set & OCR Success RE & Full RE\\
\midrule
Front & 1.15\% & 444\% \\
Angled & 0.77\% & 246\% \\
\bottomrule
\end{tabular}
\end{center}
\caption{The mean absolute relative reading error (RE) for two of the test sets. For "OCR Success" we consider examples, where OCR is successful. For "Full" we consider all examples, where a reading could be computed.}
\vspace{-0.75cm}
\label{table:reading_error_results}
\end{table}

\begin{table}
\tiny
\begin{center}
\begin{tabular}{l|cc|c}
\toprule
Image type  & Our RF & Our RE & Howells RE\\
\midrule
meter\_a & 100\% & - & 5.3\% \\
meter\_b & 0\% & 0.96\% & 7.0\% \\
meter\_c & 0\% & 1.46\% & 0.9\% \\
meter\_e & 73.2\% & 3.1\% & 1.2\% \\
meter\_f & 1.99\% & 2.5\% & 2.3\% \\
\midrule
Mean & & 2.01\% & 3.3\% \\
\bottomrule
\end{tabular}
\end{center}
\caption{The mean absolute relative reading error (RE) on Howells \etal. We also add the reading failure (RF) as the share of frames where reading failed.}
\vspace{-0.5cm}
\label{table:comparison_howells}
\end{table}

In this section, we assess our experiments using two distinct approaches. First, we measure the average absolute and relative end-to-end reading error of our system. Second, we evaluate the performance of each stage within our pipeline using various metrics and visual analysis. We explore different failure modes and limitations of our system.

\subsection{End-to-end results}

The relative reading error we consider in this section refers to the reading error expressed as the percentage relative to the entire range of the scale. When assessing reading error, we differentiate between two scenarios: whether the OCR successfully recognizes at least two scale markers correctly or fails to do so. The failure of the OCR stage can be attributed to any of the reasons outlined in section \ref{ocr_failure}.

As evident from Table \ref{table:reading_error_results}, if the scale markers are successfully recognized, our algorithm produces very accurate readings with a deviation of approximately 1\% from the correct values on average. These results highlight the potential of our algorithm in determining gauge readings if the OCR stage operates effectively.

It is evident from the results in Table \ref{table:reading_error_results} that an incorrect detection of a scale marker negatively impacts the accuracy of the readings. Strategies to mitigate such cases are discussed in detail in section \ref{ocr_failure}.

The results of the comparison we conduct between our approach and the method proposed by Howells \etal \cite{realTimeGaugeTranscription} using the dataset they provided are presented in Table \ref{table:comparison_howells}. Our approach yielded comparable or even better results than Howells et al. for most of the gauges.

For the gauge type of meter\_a our OCR model fails, which resulted in an inability to compute any reading. Additionally, for meter\_e our algorithm fails to compute a reading regularly due to issues in the OCR stage, as well as the fact that the type of notches in this meter were not represented in our training set. Consequently, the start and end point detection failed in approximately half of the cases for this meter.
It is worth emphasizing again that our approach does not require prior knowledge about the full range of the gauge's scale, unlike the method proposed by Howells \etal. This aspect makes our approach applicable to a wider range of gauges.

\subsection{Individual stage results}

\begin{table}
\tiny
\begin{center}
\begin{tabular}{l|cccc}
\toprule
& \multicolumn{4}{c}{Failure rates} \\
Test set & Gauge & Ellipse & Needle & OCR\\
\midrule
Front & 0\% & 5\% & 2.5\% & 40\% \\
Angled & 0\% & 8\% & 0\% & 57.5\%\\
Rotated & 0\% & 5\% & 5\% & 42.5\% \\
Angled \& Rotated & 0\% & 17\% & 0\% & 50\% \\
Untypical & 0\% & 50\% & 50\% & - \\
\bottomrule
\end{tabular}
\end{center}
\caption{Failure rates of different parts of the pipeline. We define a failure case of a stage, when it either heavily distorts the reading negatively or when it prevents the algorithm from computing a reading at all.}
\vspace{-0.5cm}
\label{table:failure_results}
\end{table}

\begin{table}
\tiny
\begin{center}
\begin{tabular}{l|ccc}
\toprule
Test set & Gauge IoU & Needle IoU & OCR \% \\
\midrule
Front & 0.81 & 0.80 & 33 \% \\
Angled & 0.80 & 0.72 & 18 \% \\
Angled \& Rotated & 0.81 & 0.75 & 29 \% \\
Untypical & 0.81 & 0 & 40 \% \\
\bottomrule
\end{tabular}
\end{center}
\caption{Evaluation of different stages in our pipeline. We compute the mean of the gauge detection IoU, the mean of the needle detection IoU and in the final column the mean share of the detected scale markers. IoU refers to intersection over union.}
\vspace{-0.5cm}
\label{table:stage_results}
\end{table}

\begin{table}
\tiny
\begin{center}
\begin{tabular}{l|cc}
\toprule
Test set & \% Detected & \% Non-assigned\\
\midrule
Front & 94.6 \% & 2.2 \% \\
Angled & 82.9 \% & 9.7 \% \\
Angled \& Rotated & 77.5 \% & 8.5 \% \\
Untypical & 30 \% & 58.3 \% \\
\bottomrule
\end{tabular}
\end{center}
\caption{Evaluation of the notch detection. We compute the mean share of detected annotated notches and mean share of non-assigned predicted notches. Both averages are taken over all the images in each set.}
\vspace{-0.5cm}
\label{table:keypoint_results}
\end{table}

\begin{table}
\tiny
\begin{center}
\begin{tabular}{l|ccc}
\toprule
Test set & Nothing & Persp. & Rot. \& Persp.\\
\midrule
Front & 35\% & 37\% & 33\% \\
Angled & 18\%  & 20\%  & 18\% \\
Angled \& Rotated & 0 & 0 & 29\% \\
\bottomrule
\end{tabular}
\end{center}
\caption{The mean share of detected scale markers with no corrections, with only perspective correction and with both perspective and rotation correction. }
\vspace{-0.5cm}
\label{table:ocr_ablation}
\end{table}

In Table \ref{table:failure_results} we assess the frequency of failures caused by each stage of the pipeline. We define a failure of a stage, when it either heavily distorts the reading negatively, or when it prevents the algorithm from computing a reading in the first place. For instance, if we fail to detect multiple scale markers using OCR, we subsequently cannot fit a line and therefore cannot compute a reading.

In Table \ref{table:stage_results} we evaluate the results of different parts of the pipeline. To assess the gauge detection and needle segmentation stages, we calculate the intersection over union (IoU) metric. For the OCR stage, we evaluate the performance by checking if each annotated scale marker has a predicted scale marker with an IoU of at least 0.5. If such a match exists, we define it as a detected scale marker. Otherwise, it is classified as undetected. In Table \ref{table:stage_results} the average value of the share of the detected scale markers is depicted.

In the following sections, we will conduct a detailed analysis of the performance of each individual stage. By closely examining failure cases, we aim to propose solutions to prevent such occurrences in the future.

\subsubsection{Gauge detection}

Tables \ref{table:failure_results} and \ref{table:stage_results} show the effectiveness of our gauge detection stage, as it successfully works for all the images in the test set. Additionally, the results in Table \ref{table:stage_results} demonstrate that the gauge detection remains stable even when dealing with varying perspectives or gauges that differ significantly from those in the training set.

These findings are aligned with results of previous works \cite{analogClockReading, realTimeGaugeTranscription}, who also used off-the-shelf detection models. Given that our detection model is not limited to the detection of a single gauge, it allows our system to read multiple gauges within the same image.

\subsubsection{Notch detection and ellipse fitting}

\begin{figure}[t]
\begin{center}
\includegraphics[width=0.6\linewidth]{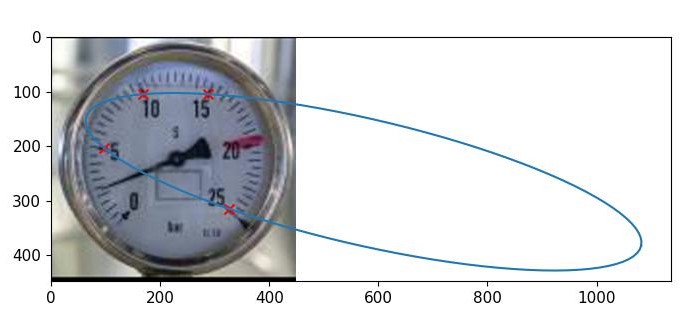}
\end{center}
\vspace{-0.5cm}
\caption{A gauge, with only four notch detections. The fitted ellipse does not fit the scale well.}
\vspace{-0.5cm}
\label{fig:ellipse_fail}
\end{figure}

Table \ref{table:keypoint_results} shows the performance of notch detection on two different metrics. The first measures the share of annotated keypoints that have at least one predicted keypoint in close proximity. The second shows the share of predicted keypoints that are close to at least one annotated keypoint. The second metric is needed to evaluate the number of wrongly predicted notches, which can disturb the ellipse fit.

We find that the notch detection and the subsequent ellipse fitting work very robustly even with different perspectives, rotations, reflections and lighting conditions. This is supported by the corresponding results presented in Table \ref{table:keypoint_results}.

From the results in Table \ref{table:keypoint_results}, we can see that the performance drops slightly for angled and rotated gauges. Furthermore, for gauges that look significantly different from the gauge types in our training data, the performance deteriorates substantially. To address this issue, a potential solution is to augment the training set with a more diverse range of gauges, such that our notch detection model generalizes better.

If our system fails to detect five or more notches, the ellipse fit will not provide any useful results, resulting in the "ellipse failure" cases reported in Table \ref{table:failure_results}. An example of such a failure case can be seen in Figure \ref{fig:ellipse_fail}.

The ellipse fitting failure could be addressed by either also predicting minor notches or segmenting pixels on the scale, and fitting the ellipse through the dense points. Another option would be to add a regularization term to the ellipse fitting which would bias the ellipse closer to a circular shape in the case of low data points. The axis of the needle could also be used as a constraint in ellipse fitting. An end-to-end differentiable approach predicting and supervising directly on an an ellipse could also be explored.

\subsubsection{Needle segmentation}

The needle segmentation stage has shown good performance for most images in the dataset across various conditions, as shown in Table \ref{table:stage_results}.

However, in our proprietary dataset, there were instances where the presence of needle shadows perturbed the segmentation result. This could be attributed to the absence of such cases in our training data. A potential solution would be to augment the training dataset with examples that include shadows, which could lead to a more robust model.

The proprietary dataset also contained gauges with a white needle on an almost white background, which our segmentation model failed to identify. These cases explain the zero IoU score in Table \ref{table:stage_results} for the needle segmentation of untypical gauges. This is likely due to a bias in our training set, which almost exclusively contains needles with good contrast between the needle and the background. Similarly, this could be addressed with a more diverse training dataset for the segmentation model.

\subsubsection{Optical character recognition} \label{ocr_failure}

\begin{figure}[t]
\begin{center}
\includegraphics[width=\linewidth]{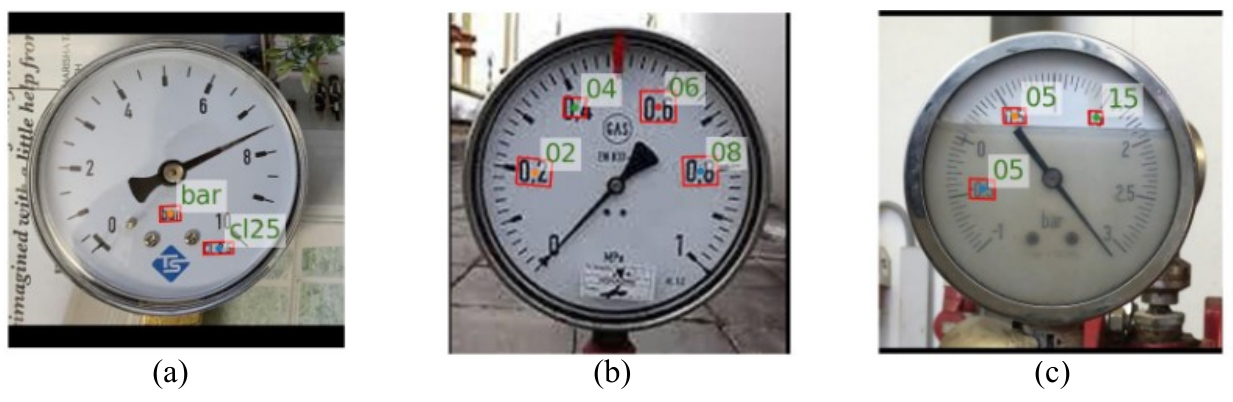}
\end{center}
\vspace{-0.5cm}
\caption{OCR failures. (a) scale markers are not recognized. (b) decimal numbers are not recognized correctly (c) negative and decimal scale markers are not recognized.}
\vspace{-0.5cm}
\label{fig:ocr_fail}
\end{figure}

Thanks to our design choice, the OCR model only needs to recognize at least two scale markers for our algorithm to predict an accurate reading. This means that even with an imperfect OCR model, we still get reasonable results.

The OCR stage of our pipeline is the most common failure mode of our system. The OCR model struggles to recognize the digits of the scale markers in many cases, even for images captured from the front perspective. This is evident from the low success rate and the limited detection of scale markers shown in Tables \ref{table:failure_results} and \ref{table:stage_results}.
Figure \ref{fig:ocr_fail}(a) shows an example of a gauge where the OCR stage fails completely. In this case, no reading can be obtained.

This reveals an insufficient performance of the pre-trained text detection model DBNet \cite{DBNet} for the scale markers on gauges. When qualitatively testing several other detection models like Mask-RCNN \cite{MaskRCNN} or DRRG \cite{DRRG}, we did not find any pre-trained model that was performing well universally among all our diverse types of gauges.

Additionally, our text recognition model ABINet \cite{ABINet} lacks the ability to detect decimal points and negative signs, limiting the scale markers we can read to natural numbers. This limitation can be observed in Figure \ref{fig:ocr_fail}(b) and (c). 

A potential solution to these issues could involve fine-tuning an existing pre-trained model specifically on the scale markers of gauges or training a text detection architecture designed specifically for scale markers and designing a language model which would be better suited for gauge reading. The text recognition model could be fine-tuned or retrained specifically with decimal and negative numbers.

The results of the ablation study are shown in Table \ref{table:ocr_ablation} and demonstrate the impact of perspective correction and rotation correction. The application of perspective correction shows a slight increase of two percent in the detection of scale markers for gauges captured from an angle. However, there is still a noticeable discrepancy between predictions obtained from the front and from an angle, suggesting that our method of perspective correction may be insufficient. A learning-based approach for perspective correction for analog clocks and by extension for analog gauges was proposed by Yang \etal \cite{analogClockReading} and could be integrated with our algorithm.

The rotation correction plays a crucial role in enabling OCR detection for rotated images, as evident from the results in Table \ref{table:ocr_ablation}. This makes it an essential step for an effective OCR detection. However, it should be noted that rotation correction slightly decreases the detection rate of scale markers in other categories where rotation is not required. This is primarily due to cases where the start and end points are inaccurately estimated, resulting in incorrect gauge orientation prediction.

\section{Conclusions}
In this paper, we presented an algorithm to read analog gauges with minimal prior information about the gauge. Our gauge reading algorithm is able to read gauges within 2\% relative precision. As shown in our experiments, the most error-prone part of our algorithm is the OCR step. Future work will focus on improving this step, possibly by fine-tuning the OCR reader on text specific to gauges and identifying helpful pre-processing steps.

Future work might also focus on building active gauge reading algorithms, where the robot reading the gauge is adjusting its viewpoint to maximize the readability of the gauge and possibly combining readings across images.

\section{Acknowledgements}
This project has received funding from the European Union's Horizon 2020 research and innovation programme under grant agreement No 871542.

{\small
\bibliographystyle{ieee_fullname}
\bibliography{bibliography}
}

\end{document}